%% file: main.tex
\documentclass{article}

\usepackage{arxiv}

\usepackage[utf8]{inputenc} 
\usepackage[T1]{fontenc}    
\usepackage{hyperref}       
\usepackage{url}            
\usepackage{booktabs}       
\usepackage{amsfonts}       
\usepackage{nicefrac}       
\usepackage{microtype}      
\usepackage{lipsum}		
\usepackage{graphicx}
\usepackage{natbib}
\usepackage{doi}
\usepackage{url}
\usepackage{footnote}
\usepackage{tablefootnote} 
\makesavenoteenv{tabular}

\usepackage[usenames,dvipsnames]{xcolor}
\usepackage{array}
\usepackage{CJKutf8}
\usepackage{balance}
\usepackage{multirow}
\usepackage{subcaption}
\usepackage{placeins}
\graphicspath{ {figs/} }
\usepackage{amsmath, amssymb}
\usepackage{mathtools} 
\usepackage{autobreak}
\usepackage{type1cm}
\usepackage{environ}
\NewEnviron{myEquation}{
\begin{equation}
\scalebox{0.9}{$\BODY$}
\end{equation}
}
\usepackage{xspace}
\usepackage{bm}

\newcommand{\eg}{\textit{e.g.}}
\newcommand{\ie}{\textit{i.e.}}
\newcommand{\methodname}{\textit{\textbf{VideoAdviser}\xspace}}

\title{VideoAdviser: Video Knowledge Distillation for Multimodal Transfer Learning}

\author{Yanan Wang${^{1,2}}$ \quad Donghuo Zeng${^1}$\quad Shinya Wada${^1}$\quad Satoshi Kurihara${^2}$ \vspace{0.3em} \\
{${^1}$KDDI Research \quad ${^2}$Keio University} \\
{{\tt \small${^1}$\{wa-yanan,do-zeng,sh-wada\}@kddi.com} \quad
{\tt \small ${^2}$\{wang-yanan,satoshi\}@keio.jp}
}
}

\date{}

\renewcommand{\headeright}{}

\begin{document}
\maketitle

\input{000_abstract}
\input{001_introduction}

\input{002_relatedwork}
\input{003_architecture}
\input{004_experiment}
\input{005_conclusion}

\bibliographystyle{unsrtnat}
\bibliography{references}  






\end{document}

%% file: 000_abstract.tex
\begin{abstract}
Multimodal transfer learning aims to transform pretrained representations of diverse modalities into a common domain space for effective multimodal fusion. However, conventional systems are typically built on the assumption that all modalities exist, and the lack of modalities always leads to poor inference performance. Furthermore, extracting pretrained embeddings for all modalities is computationally inefficient for inference.
In this work, to achieve high efficiency-performance multimodal transfer learning, we propose \methodname, a video knowledge distillation method to transfer multimodal knowledge of video-enhanced prompts from a multimodal fundamental model (teacher) to a specific modal fundamental model (student). 
With an intuition that the best learning performance comes with professional advisers and smart students,  we use a CLIP-based teacher model to provide expressive multimodal knowledge supervision signals to a RoBERTa-based student model via optimizing a step-distillation objective loss---first step: the teacher distills multimodal knowledge of video-enhanced prompts from classification logits to a regression logit---second step: the multimodal knowledge is distilled from the regression logit of the teacher to the student. 
We evaluate our method in two challenging multimodal tasks: video-level sentiment analysis (MOSI and MOSEI datasets) and audio-visual retrieval (VEGAS dataset). 
The student (requiring only the text modality as input) achieves an MAE score improvement of up to \textbf{12.3\%} for MOSI and MOSEI. Our method further enhances the state-of-the-art method by \textbf{3.4\%} mAP score for VEGAS without additional computations for inference. 
These results suggest the strengths of our method for achieving high efficiency-performance multimodal transfer learning.
\end{abstract}

%% file: 001_introduction.tex
\section{Introduction}\label{sec:intro}

\begin{figure}[!ht]
\centering
\includegraphics[keepaspectratio, scale=0.87]{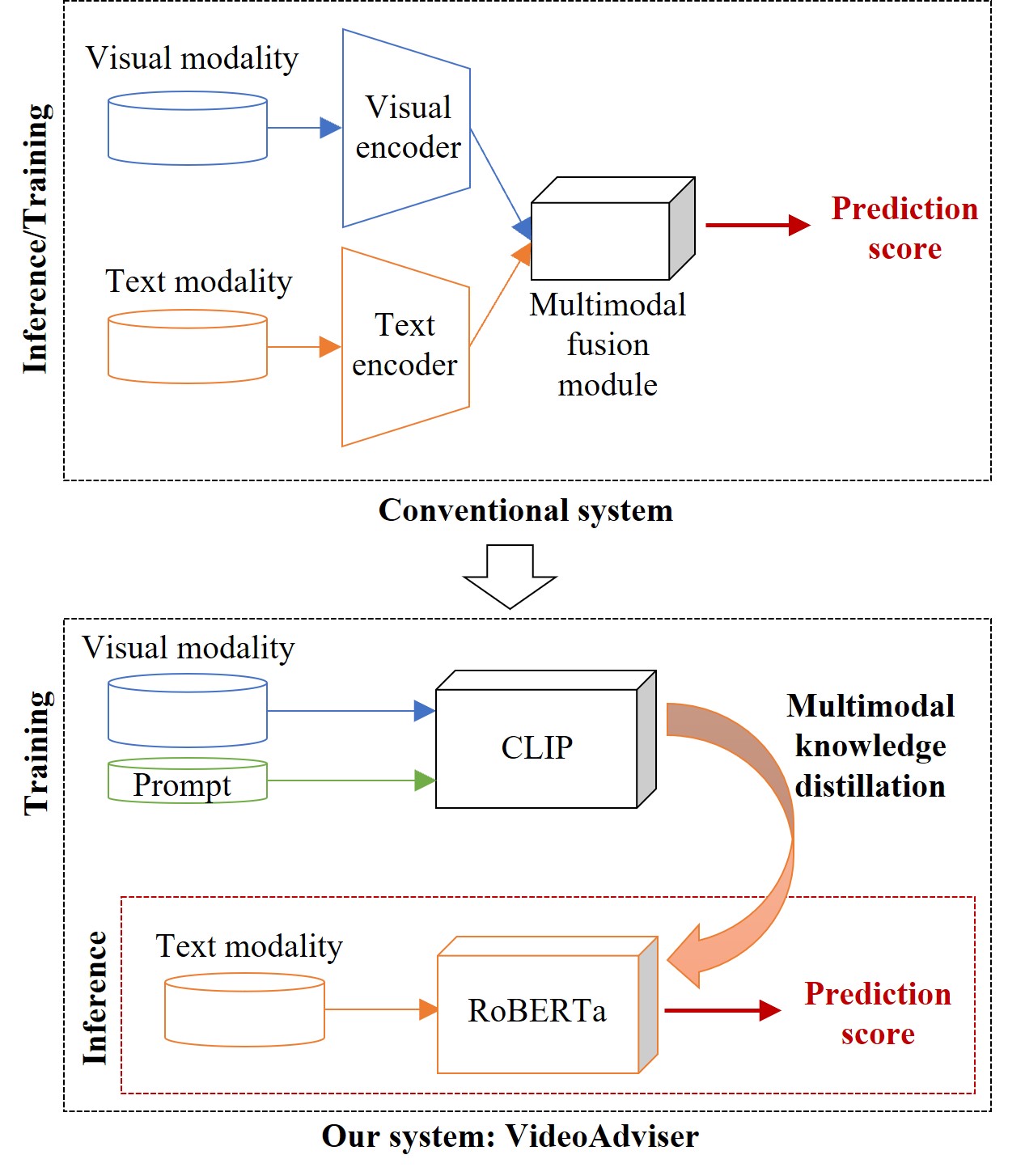}
\caption{A conceptual diagram illustrates the difference between the conventional system and our system: our system focuses on transferring multimodal knowledge from a multimodal fundamental model (\eg, CLIP) to a language fundamental model (\eg, RoBERTa-Large), and requires text only to achieve high efficiency-performance inference. On the other hand, the conventional system focuses on multimodal fusion and requires complex modules (diverse modal encoders and a multimodal fusion module) for inference.}
\label{fig:videoAdviser}
\end{figure}

Transfer learning is a promising methodology that focuses on transferring pretrained representation domains to nearby target domains~\cite{TransLearning}. For instance, finetuning a pretrained language model on a small annotated dataset enables high-performance text sentiment analysis~\cite{hazarika2020misa}.
Recent fundamental models on diverse modalities such as language models (\eg, RoBERTa~\cite{Liu2019RoBERTaAR}, GPT-3~\cite{gpt3}), visual models (\eg, ViT~\cite{dosovitskiy2020vit}), and multimodal models (\eg, CLIP~\cite{Radford2021LearningTV}, MEET~\cite{Video_Captioning}) have millions of parameters and can provide robust modal representations.
With such advancement, multimodal transfer learning aims to transform pretrained representations of diverse modalities into a common domain space for effective multimodal fusion~\cite{MultiTransLearning,MultiTransLearning_task}. It has been broadly applied to multimodal tasks such as video-level sentiment analysis~\cite{yu2021le,han-etal-2021-improving,hu-etal-2022-unimse}, and audio/text-video retrieval tasks~\cite{zeng2020deep,zhen2019deep,hou2021bicnet,ZengWWI22}.

Existing works on multimodal transfer learning unify adversarial learning to regularize the embedding distributions between different modalities, leading to effective multimodal fusion~\cite{XLYSS17,wang2017adversarial,ZhangPY18,zhen2019deep,yanan2022vae}. However, conventional systems are typically built on the assumption that all modalities exist, and the lack of modalities always leads to poor inference performance. For instance, vision-language models typically fail to achieve expected performance when given only text data as input. Furthermore, extracting pretrained embeddings for all modalities is computationally inefficient for inference. Therefore, improving robust multimodal transfer learning to achieve high efficiency-performance inference is crucial for practical applications, which motivates this work. 

Knowledge distillation (KD) is first proposed for achieving an efficient student model by transforming embedded knowledge in the predicted logits of the teacher model to a smaller student model~\cite{hinton2015distilling}. Recent works have expanded it to multimodal transfer learning by distilling mutual information from one modality to another~\cite{Gupta_2016_CVPR,kd_multimodal_task_2,jiao-etal-2020-tinybert,NEURIPS2020_6f5216f8,pan2020spatio}. However, these works always need to sacrifice the performance of the teacher model, requiring the teacher model and the student model distributed in neighboring domains (\eg, vision$\rightarrow$vison, text$\rightarrow$text). 

In this paper, with an intuition that the best learning performance comes with professional advisers and smart students, to achieve high efficiency-performance multimodal knowledge distillation, we propose \methodname~shown in Figure~\ref{fig:videoAdviser}, a video knowledge distillation method to transfer multimodal knowledge from a strong multimodal fundamental model (teacher) to a powerful specific modal fundamental model (student) via optimizing a step-distillation objective loss. 
As CLIP is a multimodal fundamental model pretrained with cross-modal contrastive learning on tremendous image-text pairs~\cite{Radford2021LearningTV}, we employ it as the teacher model to obtain multimodal knowledge of video-enhanced prompts by incorporating the video and text prompt representations. The teacher model utilizes CLIP's visual and text encoders to obtain video and text prompt embeddings without freezing the pretrained weights to preserve multimodal representation space learned by CLIP. By adapting transformer-based modules on these embeddings and extracted frame-level facial expression features, the teacher model acquires expressive multimodal knowledge of video-enhanced prompts by performing video and text prompt representations learning. To sufficiently absorb distilled multimodal knowledge from the teacher model, we employ a large-scale language model RoBERTa~\cite{Liu2019RoBERTaAR} as the student model. Since RoBERTa is a transformer-based architecture composed of huge parameters, we finetune its full parameters to leverage RoBERTa's powerful architecture to achieve high-performance student models for inference. 
In addition, we propose a step-distillation objective loss to distill coarse-fine grained multimodal knowledge to further improve the multimodal knowledge distillation. Motivated by multiscale representation learning enabling the fusion of enriched coarse-fine grained representations~\cite{li2022a,multiscale}, we consider that multitask with different target granularities allows the model to acquire representative knowledge at diverse granularities. For instance, classification encourages the model to separate the data point into multiple categorical classes representing an interval of consecutive real values to acquire knowledge at a coarse granularity. In contrast, regression enables the model to distinguish the data point into continuous real values instead of using classes to learn knowledge at a fine granularity.
To this end, in the first step, the teacher model distills multimodal knowledge of video-enhanced prompts from classification logits to a regression logit to unify knowledge at both coarse and fine granularity; In the second step, the unified multimodal knowledge is further distilled from the teacher model to the student model. 

We evaluate \methodname~in two challenging multimodal tasks: video-level sentiment analysis (MOSI and MOSEI datasets) and audio-visual retrieval (VEGAS dataset). The RoBERTa-based student model requiring only text data as input outperforms the state-of-the-art multimodal model's MAE score by \textbf{12.3\%} for MOSI and \textbf{2.4\%} for MOSEI. Our method also enhances the state-of-the-art audio-visual cross-modal model by \textbf{3.4\%} mAP score for VEGAS without additional computations for inference. Ablation studies further demonstrate that our method is able to improve the state-of-the-art method's MAE score by over \textbf{3.0\%} with almost half the parameters. These results suggest the strengths of our method for achieving high efficiency-performance multimodal transfer learning.

%% file: 002_relatedwork.tex
\section{Related work} \label{RelatedWork}
\subsection{Multimodal fundamental model}
CLIP~\cite{Radford2021LearningTV} is a multimodal fundamental model that learns transferable visual models from natural language supervision on a dataset of 400 million (image, text) pairs. It jointly trains an image encoder and a text encoder using contrasting learning objectives to obtain a joint multimodal representation space. Inspired by its remarkable zero-shot generation ability for downstream image tasks, the work~\cite{xclip} proposes XCLIP to expand pretrained CLIP on general video recognition by finetuning it on video data using a video-specific prompting module that enhances the video representation to the text representation. The work~\cite{gu2021open} utilizes a pretrained CLIP for open-vocabulary object detection by distilling visual knowledge from cropped image regions. In this work, we adapt a pretrained CLIP on distilling multimodal knowledge of video-enhanced prompts from the teacher model to the student model via a step-distillation objective loss.

\subsection{Knowledge distillation based transfer Learning}
In addition to achieving a lightweight student model by minimizing the KL divergence between the probabilistic outputs of a
teacher and student model~\cite{hinton2015distilling}, recent works on knowledge distillation focus on transferring representational knowledge from a teacher model to a student model~\cite{tian2019crd,gu2021open,wang2022kd}. For instance, these works~\cite{wang2020kd_emotion_1, wang2020kd_emotion_2} distill linguistic knowledge from a text encoder to a visual encoder by learning the mapping between modal representations. The work~\cite{teachtext} utilizes multiple text encoders to perform cross-modal knowledge distillation for stronger text-video retrieval. The work~\cite{dai-etal-2022-enabling} distills expressive text representations from a generation model to the text encoder of CLIP by minimizing text-text feature distance. However, these works mostly focus on knowledge distillation in the common modal domain or show limited performance in the cross-modal domain. In contrast, to achieve expressive knowledge distillation for multimodal transfer learning tasks, we propose a RoBERTa-based student model to improve multimodal knowledge distillation by leveraging its powerful transformer architecture.

\subsection{Video-level sentiment analysis task}
Recent works~\cite{hazarika2020misa,han-etal-2021-improving,yu2021le} on video-level sentiment analysis tasks focus on improving modality fusion. The work~\cite{wang2017adversarial} proposes VAE-Based adversarial learning method to map multimodal representations to a joint domain space for improving the modality fusion process. The work~\cite{hu-etal-2022-unimse} achieves SOTA performance on MOSI~\cite{Zadeh2016MOSIMC} and MOSEI~\cite{Zadeh2018MultimodalLA} dataset by introducing a pretrained modality fusion module that fuses multimodal representation from multi-level textual information by injecting acoustic and visual signals into a text encoder. However, all these works require preprocessed multimodal embeddings as the input which is inefficient for inference. In contrast, we employ a knowledge distillation approach that requires only one specific modality leading to efficient inference.

\subsection{Audio-visual retrieval task} 
Recent works on audio-visual retrieval tasks exploit supervised representation learning methods to generate new features across modalities in a common space~\cite{hou2021bicnet, rasiwasia2014cluster, zeng2018audio, zheng2020dual, zhen2019deep, zeng2020deep, ZengWWI22, zeng2023learning}, such that the audio-visual features can be measured directly. Inspired by the C-CCA~\cite{rasiwasia2014cluster} that aims at finding linear transformations for each modality, C-DCCA~\cite{zeng2018audio} tries to learn non-linear features in the common space by using deep learning methods. Deep learning methods by using rank loss to optimize the predicted distances, such as TNN-C-CCA~\cite{zeng2020deep}, and CCTL~\cite{ZengWWI22} models, which apply triplet losses as the objective functions to achieve better results than other CCA-variant methods. The EICS model~\cite{zeng2023learning} learns two different common spaces to capture modality-common and modality-specific features, which achieves the SOTA results so far. In this paper, we enable our method to enhance the extracted audio and visual representations of the SOTA model by distilling multimodal knowledge from a CLIP-based teacher model.  

%% file: 003_architecture.tex
\begin{figure*}[t]
\centering
\includegraphics[keepaspectratio, scale=0.75]{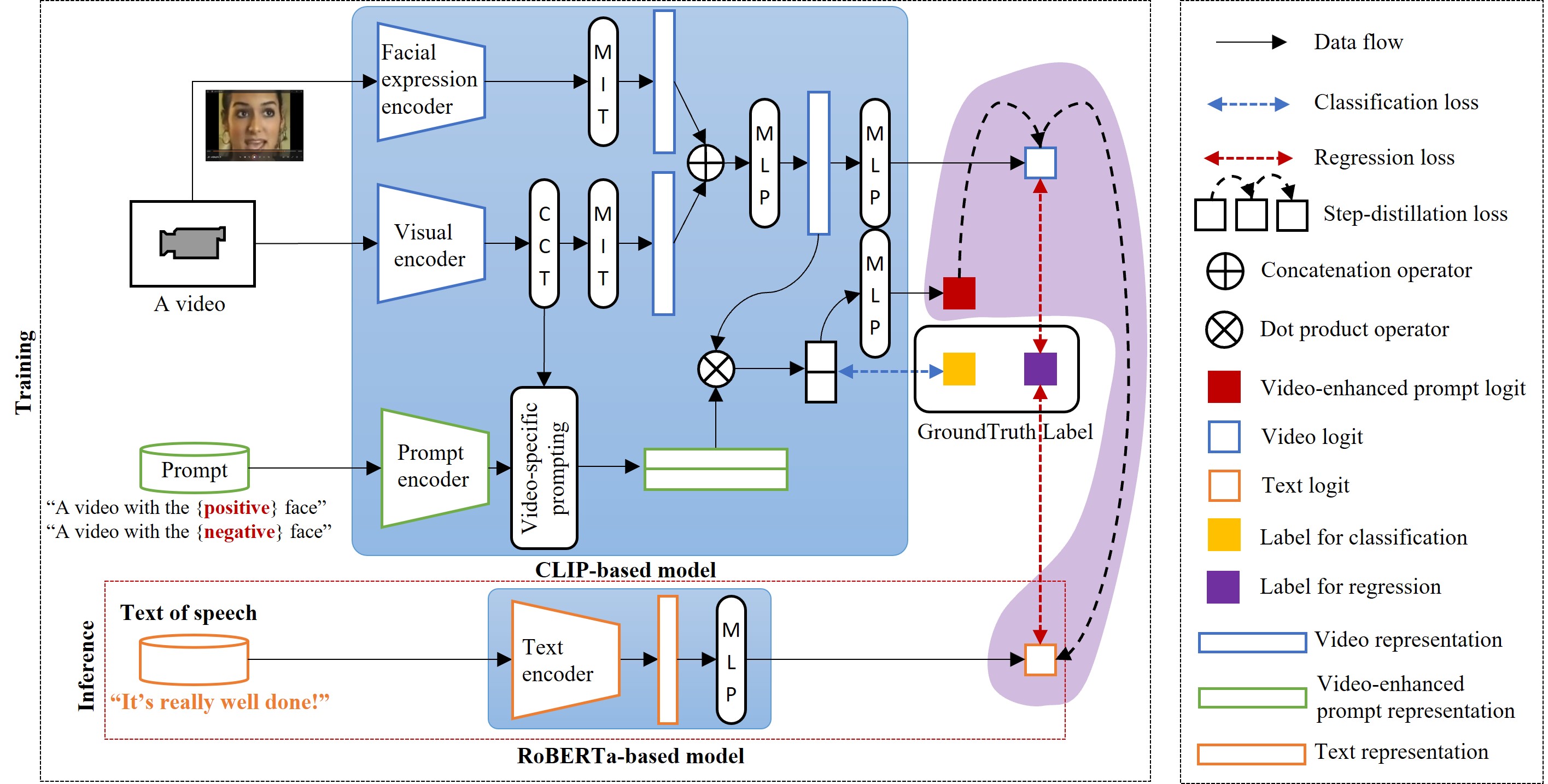}
    \caption{Architecture of \methodname~using a CLIP-based model (the teacher) to distill multimodal knowledge of video-enhanced prompts to a RoBERTa-based model (the student): the teacher model utilizes pretrained CLIP's text and visual encoders, and a facial expression encoder to obtain the sentiment class text embedding, the frame-level embedding, and the facial expression embedding. Then, the teacher model employs CCT, MIT, MLP, and a video-specific prompting module, and minimizes a binary sentiment classification loss and a sentiment regression loss. Meanwhile, the student model is finetuned on speech text by minimizing a sentiment regression loss and a step-distillation loss (the region in purple). During inference, the speech text is used to enable sentiment intensity prediction. Here, CCT, MIT, and MLP stand for the cross-frame communication transformer, multi-frame integration transformer, and multi-layer perceptron, respectively.}
\label{fig:videoAdviser_model}
\end{figure*}

\section{Problem setting}
This work focuses on video-level sentiment analysis and audio-visual retrieval tasks, respectively. For the video-level sentiment analysis task, each data point consists of a video $\bm M$, the cropped sequential face images $I$, the divided speech text $T_{speech}$, and the class text $T_{class}$, our goal is to predict the sentiment intensity $\mathcal{Z}_{pred}\in[-3,3]$ by giving only speech text $T_{speech}$ for inference.
For the audio-visual retrieval task, assume that$~\Gamma = \{\gamma_i\}_{i=1}^{N}$ is a video collection, $\gamma_i = \{a_{i}, v_{i}\}$, where $N$ indicates the data size, $a_{i}\in\mathbb{R}^{D1}$ and $v_{i}\in\mathbb{R}^{D2}$ are audio and visual features from different feature spaces. Our target aims at feeding them into a common space by mapping functions $f(x)$ and $g(x)$ to generate new features $f(a_{i})$ and $g(v_{i})$. As a result, each query $a_{i}$ for example will obtain a rank list from another modality based on $query$-$v_{j} (i\neq j)$ similarity.

\section{Methodology} \label{lab:method}
In this section, we explain our method \methodname~in detail. As shown in Fig. \ref{fig:videoAdviser_model}, our method consists of a CLIP-based model as the teacher (\S~\ref{lab:teacher}) and a RoBERTa-based model as the student (\S~\ref{lab:student}). The teacher and student models are jointly trained to achieve knowledge distillation across modalities. The student model enables sentiment intensity prediction by giving only a speech text for inference (\S~\ref{lab:training}).
We use $\mathcal{F}(\cdot)$, $\mathcal{V}(\cdot)$, $\mathcal{P}(\cdot)$ and $\mathcal{T}(\cdot)$ to denote the facial expression encoder, visual encoder, prompt encoder, and text encoder.

\subsection{The CLIP-based teacher model} \label{lab:teacher}

\paragraph{Facial expression embedding} \label{lab:fer}

To enhance the visual representations of the teacher model for sentiment intensity prediction, we first use OpenFace~\cite{openface} to crop face images $\{I_i\}^T_{i=1}\in\mathbb{R}^{P^{2}\times 3}$ with each of size $P\times P$ pixels from $T$ sampled video frames, then, we extract frame-level facial expression embedding $\bm v^{(f)}\in\mathbb{R}^{T\times D}$ with a facial expression encoder $\mathcal{F}(\cdot)$~\cite{albanie16learning} that is pretrained on the VGG-Face dataset~\cite{Parkhi15}. Here, $\bm v^{(f)}$ is an 8-dimensional sequential vector of length 64 $[T=64$, $D=8]$. More details of the pretrained model on Albanie's website \footnote{\url{https://www.robots.ox.ac.uk/~albanie/mcn-models.html}}.
\begin{equation}
  \bm v^{(f)} = \mathcal{F}(\{I_i\}^T_{i=1})
\end{equation}

\paragraph{Visual embedding}
To fully transfer the powerful generality of pretrained CLIP~\cite{Radford2021LearningTV} from image to video, we freeze the parameters of pretrained CLIP visual encoder $\mathcal{V}(\cdot)$ to obtain frame-level visual embedding $\bm v^{(v)}\in\mathbb{R}^{T\times D}$, where $T$ denotes the number of sampled video frames and $D$ is the dimension of visual embedding. Following~\cite{xclip}, given a video clip $M\in\mathbb{R}^{T \times H \times W \times 3}$ of $T$ sampled video frames with $H \times W$ pixels, we use ViT-L/14~\cite{dosovitskiy2020vit} to first divide t-th frame into $N$ patches $\{x_{t,i}\}^N_{i=1}\in\mathbb{R}^{P^{2}\times 3}$, where $t\in T$ and $N=HW/P^2$. Then, the patches $\{x_{t,i}\}^N_{i=1}$ is mapped to $\bm v^{(v)} = \{v_t^{(v)}\}^T_{t=1}$ with a linear transformation $f_m:\mathbb{R}^{P^{2}\times 3} \rightarrow \mathbb{R}^{3P^{2}\times D}$. 
\begin{equation}
  \bm v^{(v)} = \mathcal{V}(f_m (\{\bm x_{t}\}^T_{t=1}))
\end{equation}

\paragraph{Text prompt embedding}
We employ the text encoder $\mathcal{P}(\cdot)$ of pretrained CLIP to obtain text prompt embedding $\bm v^{(p)}\in\mathbb{R}^{C\times D}$ of $C$ sentiment classes by giving the sentiment class label $T_{class}\in\{$negative$, $positive$\}$, where ``positive'' class includes $0$. The text prompt such as ``A video with the $\{T_{class}\}$ face'' is generated with a text prompt generator $f_g$ and encoded as
\begin{equation}
  \bm v^{(p)} = \mathcal{P}(f_g (T_{class}))
\end{equation}

We employ the cross-frame communication transformer (CCT), multi-frame integration transformer (MIT), and video-specific prompting modules to obtain expressive multimodal sentiment knowledge. 
The CCT is a multi-layer transformer with cross-frame attention introduced in~\cite{xclip} to enable cross-frame information exchange. It is used to obtain cross-frame visual representations by giving a modified visual embedding $\bar{\bm v}^{(v)}=\{\bar{\bm v}_t^{(v)}\}^T_{t=1}$, where $\bar{\bm v}_t^{(v)}=[x_{class},v_t^{(v)}]+\bm e_{pos}$. $x_{class}$ is a learnable frame representation and $e_{pos}$ is a position embedding of patches in a frame.
The MIT is a normal transformer layer constructed by standard multi-head self-attention and feed-forward networks. Given frame-level embeddings $\bm v^{(f)}$ and $\bar{\bm v}^{(v)}$, we finally obtain the video representation $V$ as follows:
\begin{eqnarray}
& V^{(f)} = \operatorname{AvgPool}(\operatorname{MIT}(\bm v^{(f)})) \\
& V^{(v)} = \operatorname{AvgPool}(\operatorname{MIT}(\operatorname{CCT}(\bar{\bm v}^{(v)}))) \\
& V = f_v([V^{(f)}||V^{(v)}])
\end{eqnarray}
where $f_v:\mathbb{R}^{2\mathcal{D}} \rightarrow \mathbb{R}^{\mathcal{D}}$ is a two-layer MLP. $\operatorname{AvgPool}$ denotes an average pooling layer. ``$||$'' denotes a concatenation operator used to process facial expression-conditioned video representation.
We then transform the \textbf{video representation} $V$ to the \textbf{video logit} (see Fig. \ref{fig:videoAdviser_model}) with a two-layer MLP. 

Inspired by~\cite{xclip}, the teacher model employs a video-specific prompting module to enhance the prompt embedding with cross-frame visual representations. The video-specific prompting module applies a normal multi-hand attention~\cite{multi-attention} to obtain the \textbf{video-enhanced prompt representation} $\bar{\bm v}^{(p)}\in \mathbb{R}^{C\times D}$ (see Fig. \ref{fig:videoAdviser_model}) as
\begin{eqnarray}
 \bar{\bm v}^{(p)} = \bm v^{(p)} + \operatorname{Multi\_Hand\_Attention}(\operatorname{CCT}(\bm \bar{v}^{(v)}))
\end{eqnarray}

Then, we compute dot product between video representation $V$ and video-specific prompt representation $\bar{\bm v}^{(p)}=\{\bar{\bm v}_i^{(p)}\}^C_{i=1}$ to output the similarity score $\bm p = \{p_i\}^C_{i=1}$ with a softmax layer as
\begin{eqnarray}
  & p_i = \operatorname{softmax}(\bm \bar{v}^{(p)}_i\cdot V) = \displaystyle \frac{\exp(\bar{v}^{(p)}_i\cdot V)}{\sum_{i \in C}\exp(\bar{v}^{(p)}_i\cdot V)}
\end{eqnarray}
where $C$ indicates the number of sentiment classes. We further transform $\bm p$ into the \textbf{video-enhanced prompt logit} (see Fig. \ref{fig:videoAdviser_model}) with a two-layer MLP. 

\subsection{The RoBERTa-based student model} \label{lab:student}
To leverage the powerful transformer-based architecture of fundamental language models, we structure a RoBERTa-based student model~\cite{Liu2019RoBERTaAR} that consists of a text encoder $\mathcal{T}(\cdot)$ and a two-layer MLP. Given the speech text $T_{speech}$, the student model obtains text representation $V^{(t)}$ with $\mathcal{T}(\cdot)$, and output sentiment intensity $\mathcal{Z}_{pred}$ with the MLP into the \textbf{text logit} (see Fig. \ref{fig:videoAdviser_model}) as
\begin{eqnarray}
\mathcal{Z}_{pred}=\operatorname{logit}(V^{(t)}), V^{(t)} = \mathcal{T}(T_{speech})
\end{eqnarray}
Where $V^{(t)}\in\mathbb{R}^{D}$, and $\operatorname{logit}(\cdot): \mathbb{R}^{\mathcal{D}} \rightarrow \mathbb{R}^{1}$ indicates the two-layer MLP.

\subsection{Training objectives} \label{lab:training}
We simultaneously optimize the teacher and student models by applying mean squared error (MSE) loss to obtain video and text sentiment knowledge. Both teacher and student models minimize the $L_2$ distance as follows:
\begin{eqnarray}
  & \mathcal{L}^{(r)}_{v} = \scriptstyle \operatorname{MSE}(\operatorname{logit}(V), l^{(r)}) = \displaystyle \frac{1}{B}\displaystyle \sum^{B}_{i=1}{{||\operatorname{logit}(V)-l^{(r)}||}^2} \\
  & \mathcal{L}^{(r)}_{t} = \scriptstyle \operatorname{MSE}(\mathcal{Z}_{pred}, l^{(r)}) = \displaystyle \frac{1}{B}\displaystyle \sum^{B}_{i=1}{{||\mathcal{Z}_{pred}-l^{(r)}||}^2} 
\end{eqnarray}
where $B$ indicates batch size, $\mathcal{L}^{(r)}_{v}$ indicates MSE between the teacher model's video logit and sentiment label $l^{(r)}$, and $\mathcal{L}^{(r)}_{t}$ indicates MSE between the student model's text logit ($\mathcal{Z}_{pred}$) and $l^{(r)}$. Here, $\operatorname{logit}(V)$ is a 
two-layer MLP for transforming video representation $V$ into the video logit.

To learn the video-enhanced prompt representation to fuse multimodal knowledge of video and class text, we use the binary sentiment classification label $l^{(c)}$ (see Fig. \ref{fig:label_distribuation}) synthesized from the sentiment label to optimize the teacher model with a cross-entropy loss $\mathcal{L}^{(c)}_{v}$ as
\begin{eqnarray} \label{eq:12}
  & \mathcal{L}^{(c)}_{v} = -\displaystyle \sum^{C}_{i=1}{l^{(c)}_{i}\operatorname{log}(p_i)}, 
\end{eqnarray}

We optimize a step-distillation objective loss to achieve multimodal knowledge distillation from the teacher model to the student model. The step-distillation objective loss consists of a \textbf{prompt-video distance minimization} $\mathcal{L}_{p\rightarrow v}$ and a \textbf{video-text distance minimization} $\mathcal{L}_{v\rightarrow t}$, where $\mathcal{L}_{p\rightarrow v}$ is optimized to align coarse-grained classification knowledge in the video-enhanced prompt logit and fine-grained regression knowledge in the video logit, $\mathcal{L}_{v\rightarrow t}$ is optimized to align knowledge in the video logit of the teacher model and the text logit of the student model.
We apply MSE loss to perform the step-distillation as follows:
\begin{eqnarray}
  & \mathcal{L}_{p\rightarrow v} = \operatorname{MSE}(\operatorname{logit}(\bm p),\operatorname{logit}(V)) \\
  & \mathcal{L}_{v\rightarrow t} = \operatorname{MSE}(\operatorname{logit}(V), \mathcal{Z}_{pred})
\end{eqnarray}
where $\operatorname{logit}(\bm p)$ indicates the coarse‐grained classification knowledge in Eq. \ref{eq:12}.

We finally have a joint loss $\mathcal{L}$ for training the teacher and student models end-to-end as
\begin{equation}
\mathcal{L} = \alpha \mathcal{L}^{(r)}_{v} + \beta \mathcal{L}^{(r)}_{t} + \gamma \mathcal{L}^{(c)}_{v} + \delta \mathcal{L}_{p\rightarrow v} + \psi \mathcal{L}_{v\rightarrow t}
\end{equation}
where $\alpha$, $\beta$, $\gamma$, $\delta$, and $\psi$ indicate the importance of each loss value. They are empirically set as $1:10:1:10:1$ to keep all loss values on the same scale.

%% file: 004_experiment.tex
\section{Experiment}
In this section, we conducted empirical experiments on video-level sentiment analysis and audio-visual retrieval tasks to demonstrate the high efficiency-performance of our method.

\subsection{Dataset}
\begin{figure}[!h]
\centering
\includegraphics[keepaspectratio, scale=0.5]{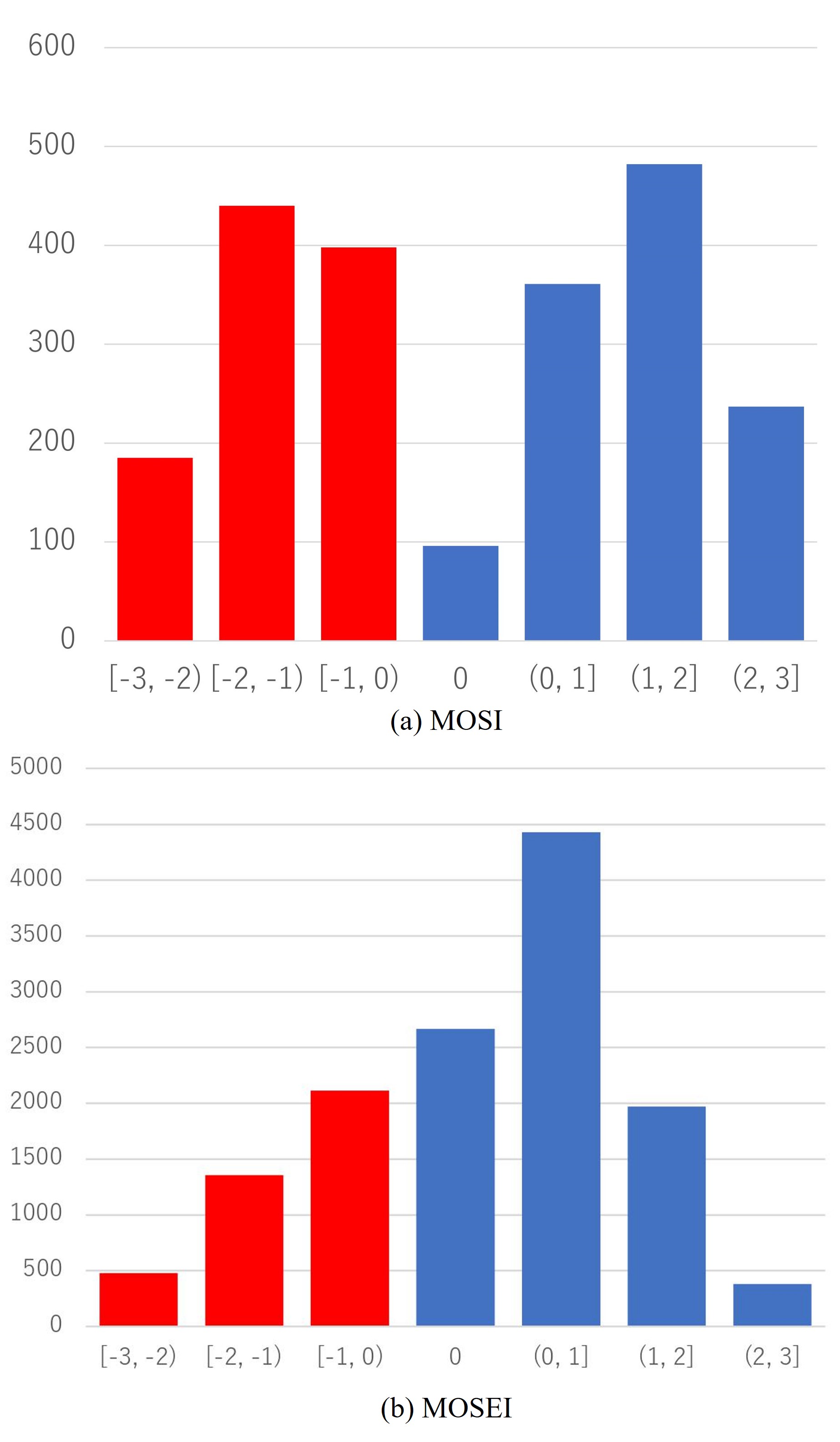}
    \caption{Label distribution of (a) MOSI and (b) MOSEI. The synthesized binary classification label is illustrated in different colors (the ``negative'' class in red color and the ``non-negative'' class in blue color).}
\label{fig:label_distribuation}
\end{figure}

MOSI~\cite{Zadeh2016MOSIMC} and MOSEI~\cite{Zadeh2018MultimodalLA} are multimodal datasets collected from online video for evaluating video-level sentiment analysis tasks. We show the dataset size in Tab.~\ref{tab:datasize}. MOSEI drops the data lacking modalities to fairly evaluate recent modality fusion-based methods~\cite{yanan2022vae}. We compared the video segment IDs of each data point for each modality and saved only the data points associated with a common segment ID. The modified MOSEI dataset was found to be more challenging than the original dataset as it lowered the strong baseline MSE score by 4.9\% (see Tab. \ref{tab:sentiment}).
Both datasets are annotated with a Likert scale in the range of $[-3,3]$, \ie, (-3: highly negative, -2: negative, -1: weakly negative, 0: neutral, +1: weakly positive, +2: positive, +3: highly positive). We further synthesize binary classification label, \ie, ([-3,0): negative, [0,3]: non-negative) used for optimizing the teacher model (\S \ref{lab:teacher}). 
The label distribution is illustrated in Fig.~\ref{fig:label_distribuation}. MOSEI is imbalanced and over $65\%$ of data is distributed in $[-1, 1]$.

VEGAS dataset~\cite{zhou2018visual} is applied for the audio-visual retrieval task, which contains 28,103 videos in total as shown in Tab. \ref{tab:datasize}. Each video can be embedded as an audio feature vector and a visual feature vector, and the audio-visual pair shares the same single label. The label represents an audio event (\eg, baby crying) of the human voice or natural sound. The number of label classes is 10, and the length of each audio-visual pair ranges from 2 to 10 seconds.


\begin{table}[h] \renewcommand{\arraystretch}{1.3} 
\caption{Dataset size. MOSEI uses the same dataset as~\cite{yanan2022vae}.}
\centering
\begin{tabular}{l|c|c|c|c}
\hline
Dataset & Train & Validation & Test & Total \\ 
\hline\hline
MOSI~\cite{Zadeh2016MOSIMC}   & 1,284 & 229 & 686 & 2,199 \\
MOSEI~\cite{Zadeh2018MultimodalLA}   & 9,473  & 1,206 & 2,710  & 13,389\\ 
VEGAS~\cite{zhou2018visual}   &22,482&- &5,621& 28,103 \\
\hline
\end{tabular}
\label{tab:datasize}
\end{table}

\subsection{Evaluation metric}
We use the mean absolute error (MAE), accuracy ($A^7$), accuracy ($A^2$), and weight-$F1$ score for evaluating MOSI and MOSEI. $A^7$ denotes a 7-class and $A^2$ denotes a binary accuracy metric. Since MOSI and MOSEI are regression problems, we consider MAE to be the most reasonable metric for fair evaluations. In addition to the binary accuracy reported by most of the previous works, we evaluate the 7-class accuracy as did the SOTA method~\cite{hu-etal-2022-unimse} to eliminate the effect of the data imbalance. For the audio-visual retrieval task, we apply the mean average precision (mAP) as previous works~\cite{zeng2023learning, ZengWWI22} to evaluate our model.


\subsection{Training setting}
\begin{figure}[h]
\centering
\includegraphics[keepaspectratio, scale=0.7]{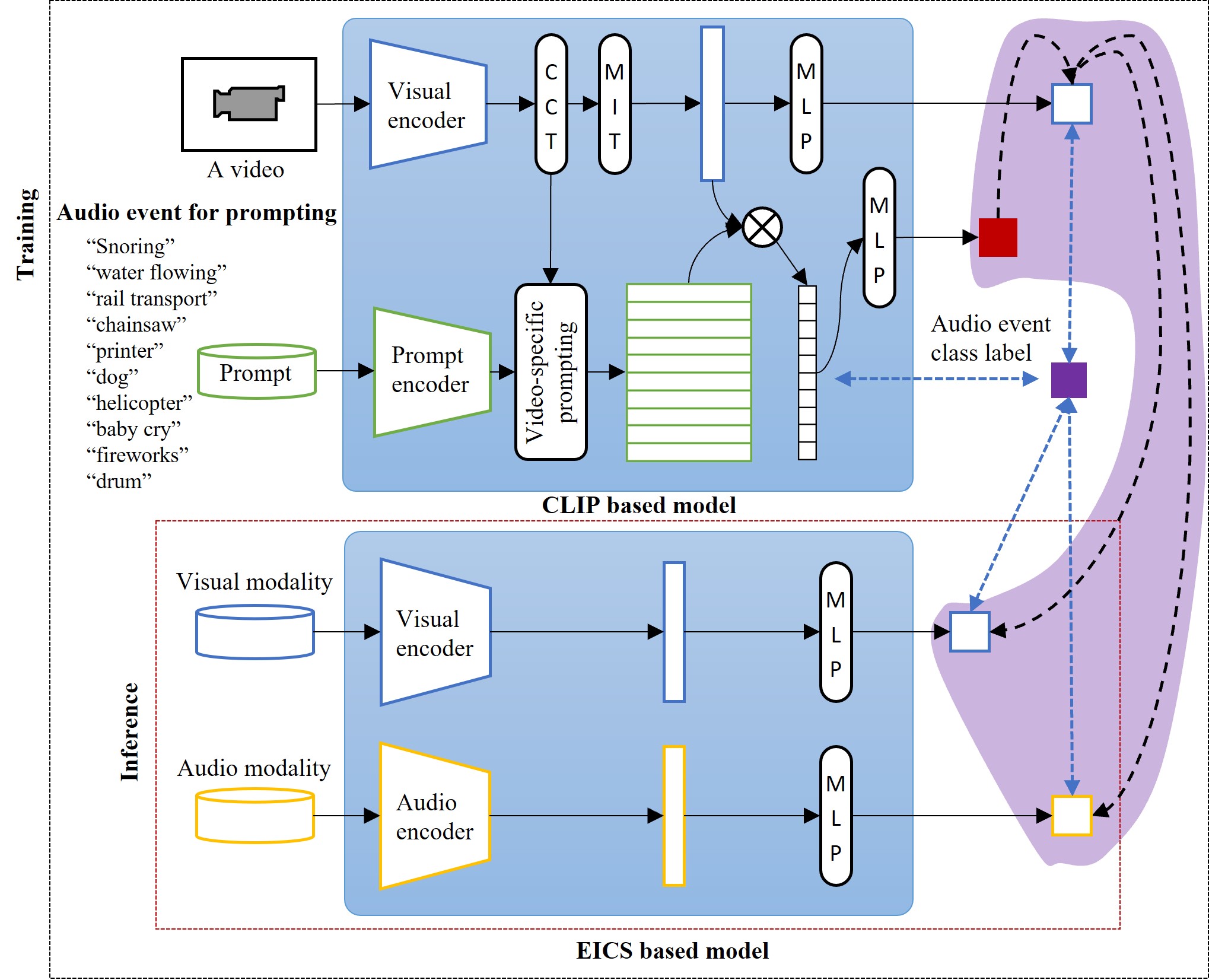}
    \caption{Architecture of \methodname~ for audio-visual retrieval task using a CLIP-based model (the teacher) to distill multimodal knowledge of video-enhanced prompts to an EICS-based audio-visual model (the student). The teacher model is finetuned for the audio event classification to distill multimodal knowledge to the student model via the step-distillation loss (the region in purple). We adopt 3-layer MLP with 128-dimensional hidden layers.}
\label{fig:audiovisual_model}
\end{figure}

We train the teacher and the student models simultaneously and use only the student model for inference. The text modality is used for evaluating MOSI and MOSEI. On the other hand, as shown in Fig.~\ref{fig:audiovisual_model}, we utilize the teacher model to distill multimodal knowledge for both visual and audio encoders of the state-of-the-art model EICS~\cite{zeng2023learning} for audio-visual retrieval tasks. Both visual and audio encoders are used as student models to evaluate VEGAS. 
We show the hyperparameters of \methodname~(\S \ref{lab:method}) for both tasks in detail in Tab.~\ref{tab:hyperparameter}. 

\begin{table}[!t] \renewcommand{\arraystretch}{1.3} 
\caption{Comparison results for MOSI and MOSEI. Our model reduces the state-of-the-art UniMSE's MAE score by {\bfseries 12.3\%} for MOSI, and VAE-AMDT's MAE by {\bfseries 2.4\%} for MOSEI. Here, ($\downarrow$) indicates the lower the MAE, the better the performance, and ($\uparrow$) indicates the vice-versa. (*) indicates the results produced on the modified MOSEI dataset.} \label{table_example}
\centering \resizebox{\linewidth}{!}{%
\begin{tabular}{l|c|c|c|c|c||c|c|c|c|c}
\hline
\multirow{2}{*}{\bfseries Model} & \multicolumn{5}{c||}{ \bfseries MOSI} & \multicolumn{5}{c}{\bfseries MOSEI} \\
\cline{2-11}
    & MAE $\downarrow$ & $A^7 \uparrow$ &$A^2 \uparrow$ &  F1 $\uparrow$ & Corr $\uparrow$ & MAE $\downarrow$ & $A^7 \uparrow$ & $A^2 \uparrow$ &  F1 $\uparrow$ & Corr $\uparrow$ \\
\hline\hline
MISA \cite{hazarika2020misa}                   & 0.804 &-& 80.8 & 80.8 &0.764& 0.568  &-& 82.6  & 82.7 &0.717\\
VAE-AMDT \cite{yanan2022vae}                    & 0.716 &-& 84.3 & 84.2 &-& 0.526* &-& 82.8*  & \bfseries 87.5*&- \\
MAG-BERT \cite{rahman-etal-2020-integrating}    & 0.712 &-& 84.2 & 84.1 &0.796& 0.539  &-& 84.7  & 84.5 &-\\
Self-MM \cite{yu2021le}                         & 0.713 &-& 84.0 & 84.4 &0.798& 0.530/0.579*  &-& 82.8/84.6*  & 82.5/84.6* &0.765/- \\
MMM \cite{han-etal-2021-improving}                  & 0.700 &46.7& 84.1 & 84.0 &0.800& 0.526  &54.2& 82.2  & 82.7 &0.772\\
UniMSE \cite{hu-etal-2022-unimse}                   & 0.691 &48.7& 85.9  & 85.3 & 0.809 & 0.523 & 54.4 & \bfseries 85.9  & 85.8& 0.773\\
\bfseries \methodname~ (ours) & \bfseries 0.568 & \bfseries 51.3 & \bfseries 87.7 & \bfseries 87.9 & \bfseries 0.872 & \bfseries 0.502*  &\bfseries54.5*& 84.5* &  85.0* &\bfseries0.810* \\
\hline
Human            & 0.710 &-& 85.7 & 87.5 &0.820& -    &  - & -   & - & - \\
\hline
\end{tabular}
}
\label{tab:sentiment}
\end{table}


\begin{table}[h] \renewcommand{\arraystretch}{1.3} 
\caption{The hyperparameters for training \methodname. Here, ``B'' denotes the batch size, ``Audio logit'' denotes the output of the audio encoder for VEGAS (see Fig. \ref{fig:audiovisual_model}).}
\centering
\begin{tabular}{c|l|c|c}
\hline
  & Hyperparameter & MOSI, MOSEI & VEGAS \\ \hline \hline

\multirow{5}{*}{\rotatebox[origin=c]{90}{Video}}  
& visual encoder & \multicolumn{2}{c}{ViT-L/14} \\
& Num. of frames & \multicolumn{2}{c}{8}  \\
& Frame size & \multicolumn{2}{c}{224$\times$224} \\
& visual embedding size (input) & {(B, 64, 8)} & (B, 1, 10) \\
& Visual hidden layer size & \multicolumn{2}{c}{(B, 128)} \\

\hline
\multirow{3}{*}{\rotatebox[origin=c]{90}{Prompt}} 
& Prompt encoder & \multicolumn{2}{c}{ClipTextModel} \\
& Prompt embedding size (input) & \multicolumn{2}{c}{(B,77,512)} \\
& Prompt hidden layer size & \multicolumn{2}{c}{128} \\
\hline

\multirow{3}{*}{\rotatebox[origin=c]{90}{Text}} 
& Text encoder & RoBERTa-large & - \\
& Text embedding size (input) &(B,100,1024) & - \\
& Text hidden layer size &128& - \\
\hline

\multirow{3}{*}{\rotatebox[origin=c]{90}{Audio}} 
& Audio encoder & - & EICS model \\
& Audio feature size (input) & - & 10 \\
& audio hidden layer size & -& 128  \\
 \hline

\multirow{4}{*}{\rotatebox[origin=c]{90}{Output logit}} 
& Video-enhanced prompt logit & (B, 1) & (B, 10) \\
& Video logit & (B, 1) & (B, 10) \\
& Text logit & (B, 1) & -  \\
& Audio logit & - & (B, 10) \\
 \hline

\multirow{4}{*}{\rotatebox[origin=c]{90}{Optimizer}} 
& Method & \multicolumn{2}{c}{AdamW~\cite{Kingma2015AdamAM}}\\
& Learning rate & \multicolumn{2}{c}{$8$e-$6$}  \\
& Warmup steps & \multicolumn{2}{c}{15}\\
& Schedular & \multicolumn{2}{c}{cosine\_schedule\_with\_warmup} \\ \hline

\multirow{3}{*}{\rotatebox[origin=c]{90}{Training}} 
& GPU & \multicolumn{2}{c}{GTX 1080 Ti} \\
& Batch size & \multicolumn{2}{c}{4}  \\
& Training epochs &\multicolumn{2}{c}{100} \\
 \hline
\end{tabular}

\label{tab:hyperparameter}
\end{table}

\subsection{Performance} \label{performance}
\subsubsection{Evaluation of video-level sentiment analysis}
We compared \methodname~with strong baseline methods on the test set of MOSI and MOSEI in Tab. \ref{tab:sentiment}. Compared with the state-of-the-art method UniMSE~\cite{hu-etal-2022-unimse} that utilizes the powerful architecture of a large-scale pretraining model T5~\cite{t5} to improve the multimodal fusion by embedding multimodal signals into an auxiliary layer of T5, \methodname~is a multimodal knowledge distillation-based method that distills multimodal knowledge from a multimodal fundamental model CLIP~\cite{xclip} to a language model RoBERTa~\cite{Liu2019RoBERTaAR}. UniMSE was trained by integrating the training datasets of MOSI, MOSEI, MELD~\cite{poria-etal-2019-meld}, IEMOCAP~\cite{Busso2008IEMOCAPIE} and multimodal signals are required for inference. In contrast, our method was trained using the target dataset and requires only text data for inference. \methodname~significantly improves UniMSE's MAE score by \textbf{12.3\%} for MOSI, and outperforms a strong baseline method VAE-AMDT's MAE score by \textbf{2.4\%} for MOSEI. As we use the teacher model to offer auxiliary multimodal supervision signals to the student model, by leveraging the strengths of the learned multimodal space of the teacher model and the large-scale parameters of the student model, we think our method is effective for achieving high-performance multimodal knowledge distillation via minimizing the step-distillation objective loss (\S \ref{lab:training}).

\subsubsection{Evaluation of audio-visual retrieval}
We further evaluated our \methodname~on the VEGAS dataset in Tab. \ref{tab:vegas_result}. Compared to the state-of-the-art method EICS~\cite{zeng2023learning} that builds two different common spaces to learn the modality-common and modality-specific features, which achieves an average mAP of 0.788. Our method utilizes the distilled multimodal knowledge to enhance the performance of EICS. As a result, it achieves an average mAP of 0.822 and improves EICS~\cite{zeng2023learning} by \textbf{3.4\%}, suggesting the generality of our method on audio-visual retrieval tasks.

\begin{table}[h]\renewcommand{\arraystretch}{1.3} 
\caption{The mAP comparison results with state-of-the-art models for VEGAS. Here, ``V'' and ``A'' indicate ``Video'' and ``Audio'', respectively.}
\begin{center}
\begin{tabular}{l|c|c|c}
    \hline
   \multirow{2}{*}{\textbf{Model}} & \multicolumn{3}{c}{\textbf{VEGAS}}  \\
    \cline{2-4}
  & A$\rightarrow$V
  & V$\rightarrow$A
  & Average
   \tabularnewline 
\hline \hline
   Random & 0.110 & 0.109 & 0.109 \\
    \hline
    BiC-Net~\cite{hou2021bicnet} & 0.680 & 0.653 & 0.667 \\ 
    C-CCA~\cite{rasiwasia2014cluster} & 0.711 & 0.704  & 0.708  \\ 
    C-DCCA~\cite{yu2018category} & 0.722 & 0.716  & 0.719  \\ 
    DCIL~\cite{zheng2020dual}  & 0.726 & 0.722 & 0.724 \\ 
    DSCMR~\cite{zhen2019deep} & 0.732 & 0.721 & 0.727 \\ 
    TNN-C-CCA~\cite{zeng2020deep} & 0.751 & 0.738 & 0.745 \\
    CCTL~\cite{ZengWWI22}  & 0.766 & 0.765 & 0.766 \\ 
    EICS~\cite{zeng2023learning} &0.797 & 0.779 &0.788 \\
    \hline
   \bfseries \methodname~(ours) & \bfseries0.825  & \bfseries0.819  &\bfseries0.822 \\
\hline
\end{tabular}
\label{tab:vegas_result}
\end{center}
\end{table}

\subsection{Efficiency} \label{efficiency}
By comparing the number of parameters with state-of-the-art models in Fig. \ref{tab:efficiency}, our proposed \methodname~requires only a language model as the student that is able to achieve a high efficiency-performance model for inference. The Student (BERT~\cite{devlin-etal-2019-bert}) achieved a compatible MAE score with fewer parameters than previous BERT-based models. Moreover, these models always process visual and audio signals for multimodal fusion, which might require more parameters and increase the computation cost. Compared with the state-of-the-art model UniMSE that uses a pretrained transformer-based language model T5~\cite{t5} to perform multimodal fusion, our model, the student (ROBERTa-Base~\cite{Liu2019RoBERTaAR}) with nearly half of the parameters reduces MAE score of over \textbf{3.0} point, suggesting the high efficiency-performance of our method. \methodname~was further improved over \textbf{9.0} point by adopting a RoBERTa-Large model as the student model.   

\begin{table}[!h] \renewcommand{\arraystretch}{1.3} 
\caption{Efficiency comparison. \methodname~is able to train a high efficiency-performance student model compared to state-of-the-art methods for inference. The student (RoBERTa-Base) outperforms the SOTA by over \textbf{3.0} point with nearly half the parameters.}
\centering 
\begin{tabular}{l|c|c}
\hline
\multirow{2}{*}{\textbf{Model}} & \multirow{2}{*}{\textbf{Parameters}}  & \multicolumn{1}{c}{\textbf{MOSI}} \\
\cline{3-3}
&&  MAE  \\

\hline\hline
\bfseries BERT-based model &&\\
- MISA~\cite{hazarika2020misa}     & >110M & 0.804  \\
- MAG-BERT~\cite{rahman-etal-2020-integrating} & >110M & 0.712  \\
- Self-MM~\cite{yu2021le}  & >110M & 0.713  \\
- MMM~\cite{han-etal-2021-improving}  & >110M & 0.700  \\
\hline
\bfseries T5-based model &&\\
- UniMSE~\cite{hu-etal-2022-unimse}    & >231M & 0.691 \\
\hline
\bfseries RoBERTa-based model &&\\
- VAE-AMDT~\cite{yanan2022vae}  & >355M & 0.716 \\
\hline
 \bfseries \methodname~(ours) &&\\
- Student (BERT) & \bfseries 110M & 0.704\\
- Student (RoBERTa-Base) & 125M & 0.660 \\
- Student (RoBERTa-Large) & 361M & \bfseries 0.568 \\
\hline
\end{tabular}
\label{tab:efficiency}
\end{table}

\subsection{Analysis} \label{ablation}

\begin{figure*}[!t]
\centering
\includegraphics[keepaspectratio, scale=0.32]{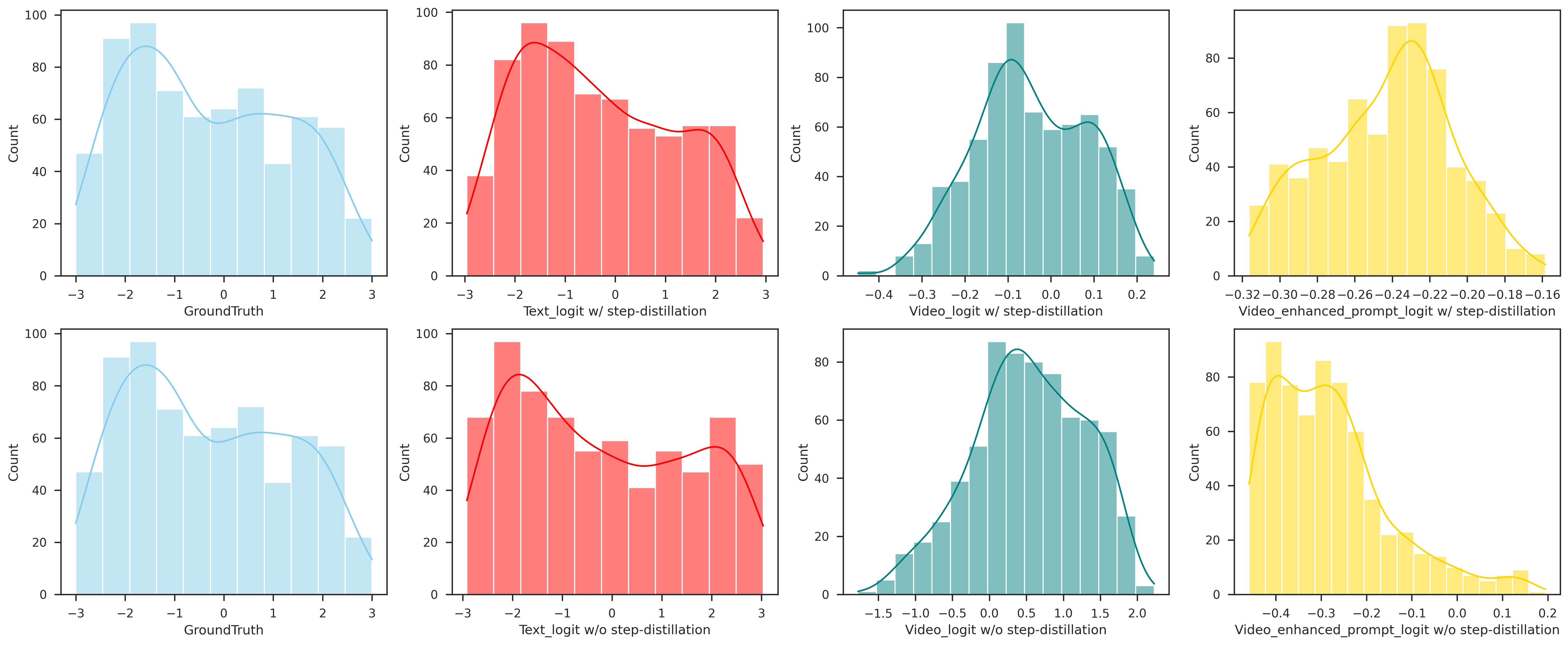}
\caption{Visualization of logistic knowledge distribution with and without the step-distillation objective loss. The top row plots the histograms of logit by applying the step-distillation, and the bottom row indicates the vice-versa. The groudTruth indicates the label distribution, and text\_logit indicates the predicted regression score of the student model. Our method using the step-distillation (the top) demonstrates a distribution of regression scores close to the groundTruth, affected by the knowledge distribution of the ``video\_logit'' and ``video\_enhanced\_prompt\_logit''.}
\label{fig:step_distill}
\end{figure*}

\subsubsection{Effectiveness of components of the teacher model}
We studied the effects of two core components of the teacher model (Facial expression encoder and video-specific prompting module) in Tab. \ref{tab:component}. The results show that these two components help improve the multimodal knowledge distillation and boost the final performance of the student model. We believe that the facial expression encoder provided extra visual knowledge, and the video-specific prompting module further associated visual knowledge with text prompt representations encoded by the prompt encoder.

\begin{table}[h] \renewcommand{\arraystretch}{1.3} 
\caption{Ablation results show the effects of components of the teacher model for multimodal knowledge distillation on MOSI dataset.}
\centering 
\begin{tabular}{l|c|c|c|c}
\hline
\multirow{2}{*}{\bfseries Model} & \multicolumn{4}{c}{ \bfseries MOSI} \\
\cline{2-5}
    & MAE & $A^7$ & $A^2$ &  F1 \\
\hline\hline
\methodname~(ours) &\bfseries 0.568 & \bfseries 51.3 & 87.7 &  \bfseries 87.9 \\
- w/o Facial expression encoder & 0.579 &  50.2 &  86.8 &  86.4 \\
- w/o Video-specific prompting & 0.570 & 50.1 & \bfseries 88.1 &  87.7 \\

\hline
\end{tabular}
\label{tab:component}
\end{table}

\subsubsection{Effectiveness of the student model}
We studied the effects of \methodname~on different student models in Tab. \ref{tab:student}. We select two language models (BERT and RoBERTa) that have frequently been used in recent works~\cite{hazarika2020misa,yanan2022vae,rahman-etal-2020-integrating,yu2021le,han-etal-2021-improving}. By comparing the performance of language models with and without adopting a teacher model, the results demonstrate that our method improves a general language model's MAE score by over \textbf{6.0} point on average, suggesting the efficacy and generality of our method with different student models. We consider that the teacher model offers auxiliary multimodal supervision to the student model during training, the language model-based students are able to learn multimodal knowledge from the teacher with their large-scale parameters.

We further trained a student model by freezing pretrained parameters, which dramatically dropped the MAE score from $0.568$ to $1.478$. This result makes us believe that in order to achieve expressive multimodal knowledge distillation across modalities, it is essential to finetune full parameters to leverage the strengths of large-scale pretrained models with powerful representational learning capabilities.

\begin{table}[!h] \renewcommand{\arraystretch}{1.3} 
\caption{Effects in different student models. Our method improves the MAE score of pretrained language models by over \textbf{6.0} point on average.}
\centering 
\begin{tabular}{l|c|c|c}
\hline
\multirow{2}{*}{\bfseries Model} & \multicolumn{3}{c}{ \bfseries MOSI} \\
\cline{2-4}
    & MAE & $A^2$ &  F1 \\
\hline\hline
Teacher (CLIP-based model)     & - & 57.3 & - \\
\hline
BERT w/o teacher & 0.753 & 84.1 & 83.6   \\
Student (BERT) & 0.704 & 84.7 & 83.8   \\
\hline
RoBERTa-Base w/o teacher  & 0.719 & 84.6 & 84.3  \\
Student (RoBERTa-Base)  & 0.660 & 85.4 & 84.6  \\
\hline
RoBERTa-Large w/o teacher  & 0.660 & 87.3 & 87.3 \\
\bfseries Student (RoBERTa-Large) & \bfseries 0.568 & \bfseries 87.7 &  \bfseries 87.9 \\
\hline
\end{tabular}
\label{tab:student}
\end{table}

\subsubsection{Modality effectiveness}
To confirm the robustness of \methodname~in multimodal knowledge distillation not only for text modality but also for diverse modalities such as visual and audio modalities, we respectively studied the effects on visual and audio modalities for audio-visual retrieval tasks. As the results indicated in Tab. \ref{tab:vegas_ablation}, the proposed step-distillation works for both modalities by boosting the baseline EICS model by over 1\% mAP score. By associating both sides, we finally improved the baseline by 3.4\%.

\begin{table}[h] \renewcommand{\arraystretch}{1.3} 
\caption{Ablation results show the effects of step-distillation on audio and video modalities for VEGAS. Here, ``w/ video distillation'' indicates that the step-distillation is only adopted for the visual modality of the student model, ``w/ audio distillation'' indicates the other side, and ``w/ audio and video distillation'' indicates both sides (see Fig. \ref{fig:audiovisual_model}).}
\begin{center}
\begin{tabular}{l|c|c|c}
    \hline
   \multirow{2}{*}{\textbf{Model}} & \multicolumn{3}{c}{\textbf{VEGAS}}  \\
    \cline{2-4}
  & A$\rightarrow$V
  & V$\rightarrow$A
  & Average \\
\hline \hline
   baseline (EICS~\cite{zeng2023learning}) &0.797 & 0.779 &0.788\\
   \hline
   \bfseries \methodname~(ours) & & &\\
   -w/ video distillation & 0.794 & 0.810 & 0.802\\
   -w/ audio distillation & 0.791 & 0.815 & 0.803\\
   -w/ (audio and video) distillation & \bfseries0.825  & \bfseries0.819  &\bfseries0.822 \\
   
\hline
\end{tabular}
\label{tab:vegas_ablation}
\end{center}
\end{table}

\subsubsection{Effectiveness of dataset size}
In general, the larger the dataset, the better the performance. We trained \methodname~with a combination of the MOSI and MOSEI datasets to see if we can further improve the performance. As the results indicated in Tab. \ref{tab:dataset_ablation}, The model performs much better than those trained on individual datasets and suggests the efficacy of our approach for different dataset sizes.

\begin{table}[h] \renewcommand{\arraystretch}{1.3} 
\caption{Results of \methodname~trained with a combination of MOSI and MOSEI datasets. The model performs much better for both the MOSI and MOSEI test sets. Here, (*) denotes the result of the model trained on the individual dataset.}
\centering 
\begin{tabular}{l|c|c|c}
\hline
\bfseries Test dataset & MAE & $A^7$ & $A^2$  \\
\hline\hline
MOSI & \textbf{0.546} (0.568) & \textbf{51.3} (51.3) & \textbf{88.5} (87.7) \\
MOSEI & \textbf{0.491} (0.502) & \textbf{55.6} (54.5) & 84.2 (\textbf{84.5}) \\
MOSI+MOSEI  & 0.502 & 54.79 & 85.05 \\

\hline
\end{tabular}
\label{tab:dataset_ablation}
\end{table}

\begin{table}[h] \renewcommand{\arraystretch}{1.3} 
\caption{Ablation results show the effects of the proposed step-distillation loss for MOSI.}
\centering 
\begin{tabular}{l|c|c|c|c}
\hline
\multirow{2}{*}{\bfseries Model} & \multicolumn{4}{c}{ \bfseries MOSI} \\
\cline{2-5}
    & MAE & $A^7$ & $A^2$ &  F1 \\
\hline\hline
CRD~\cite{tian2019crd} & 0.617 & 48.8 & 86.3 & 85.9 \\
\hline
\bfseries \methodname~(ours) &&&\\
- w/o step-distillation & 0.660 & 45.5 & 87.3 & 87.3  \\
- w/o step-distillation:step1 &0.618 & 49.0& 86.5 & 86.3  \\
- w/ step-distillation  & \bfseries 0.568 & \bfseries 51.3 & 87.7 &  \bfseries 87.9 \\
\hline
\end{tabular}
\label{tab:disll_loss}
\end{table}

\subsubsection{Effectiveness of the step-distillation loss}
We ablatively studied the effects of our proposed step-distillation loss for multimodal knowledge distillation in Tab. \ref{tab:disll_loss}. Without the first step---distilling multimodal knowledge from a video-enhanced prompt logit to a video logit (see Fig. \ref{fig:videoAdviser_model}), the learned multimodal space of CLIP cannot be passed to the student model via the video logit, resulting poor student model performance. On the other hand, it improves the regular language model (w/o step-distillation) \textbf{4.2\%} MAE score and suggests the effectiveness of the second step---distilling the knowledge of the video logit from the teacher model to the student model. Moreover, by optimizing the first and second steps, our proposed method outperforms a cutting-edge contrastive representation distillation method (CRD)~\cite{tian2019crd} that proposed a contrastive-based objective for transferring knowledge between deep networks. Compared to the CRD which is designed to model mutual information across dimensions of the knowledge representations, Our proposed step-distillation applies MSE to mapping mutual information across modalities via one-dimensional logits (\ie, video-enhanced prompt logit, video logit, and text logit). Our method performs better than CRD in transferring regression information for multimodal knowledge distillation.

In addition, we show comparison results of the proposed step-distillation loss with three widely-known distillation function KD~\cite{hinton2015distilling}, FitNet~\cite{romero2014fitnets} and PKT~\cite{pkt_eccv} in Tab. \ref{tab:disll_loss_vegas}. KD and PKT are proposed to minimize the KL divergence between the probabilistic outputs of a teacher and student model. On the other hand, FitNet and our step-distillation aim at minimizing the $L_2$ distance for knowledge distillation. Compared to KD, FitNet and PKT are one-step distillation loss functions, whereas our step-distillation performs two-step distillation, with the aim of transferring multimodal knowledge across multiple scales. To achieve a fair comparison, we adapted these three approaches to our problem setting of two-step distillation. As the results indicated in Tab. \ref{tab:disll_loss_vegas}, the step-distillation outperforms other approaches and suggests its efficacy on multimodal knowledge distillation. We noted that the PKT-based two-step distillation achieves a compatible score with ours. We consider that audio-visual tasks focus on distilling multimodal knowledge of categorical audio events rather than fine-grained regressional knowledge so that transferring probabilistic knowledge of each category can also work well. Compared to KD which utilized the softmax function to obtain probabilistic knowledge, PKT adopted the cos-similarity function to better obtain dimension-level correlation with the probabilistic knowledge.

\begin{table}[h] \renewcommand{\arraystretch}{1.3} 
\caption{Comparison results between widely-known knowledge distillation loss and the proposed step-distillation loss for VEGAS.}
\centering 
\begin{tabular}{l|c|c|c}
    \hline
   \multirow{2}{*}{\textbf{Model}} & \multicolumn{3}{c}{\textbf{VEGAS}}  \\
    \cline{2-4}
  & A$\rightarrow$V
  & V$\rightarrow$A
  & Average \\
\hline \hline
   KD~\cite{hinton2015distilling} &0.783&0.612&0.701 \\
   FitNet~\cite{romero2014fitnets} & 0.803 & 0.781 &0.792 \\
   PKT~\cite{pkt_eccv} &0.824&0.807&0.816\\
   step-distillation (ours) & \bfseries 0.825 & \bfseries 0.819 & \bfseries 0.822 \\
\hline
\end{tabular}
\label{tab:disll_loss_vegas}
\end{table}

We further illustrate the logistic knowledge distribution with and without the step-distillation loss in Fig. \ref{fig:step_distill}. Compared to the ``Text\_logit w/o step-distillation'' that plots the histogram of regression scores without performing the step-distillation, ``Text\_logit w/ step-distillation'' is close to the groundTruth label distribution. Especially the distribution in the range of $[-1,1]$ is strongly affected by the teacher model. Because the ``Video\_logit w/o step-distillation'' distributes in the range of $[-1.5,2]$ and the ``Video\_enhanced\_prompt\_logit w/o step-distillation'' distributes in the range of $[-0.4,0.2]$, by performing the step-distillation, the predicted regression score produced by the student model can be affected by the gap of these different distributions, and demonstrate that our proposed step-distillation is effective for multimodal knowledge distillation.

\subsection{Significance Testing} \label{Significance_test}
We tested the stability of the performance improvement by \methodname~ using the Almost Stochastic Order test (ASO) \cite{del2018optimal, dror2019deep} as implemented by \cite{ulmer2022deep}. We compared three models, \methodname~(ours), \methodname~ w/o step-distillation (baseline), and CRD based on five random seeds each using ASO with a confidence level of $\alpha = 0.05$. ASO computes a score ($\epsilon_\text{min}$) indicated in Tab. \ref{tab:st} to represent how far the first model is from being significantly better with respect to the second. $\epsilon_\text{min} = 0$ represents truly stochastic dominance and $\epsilon_\text{min} < 0.5$ represents almost stochastic dominance. 

\begin{table}[!h] \renewcommand{\arraystretch}{1.3} 
\caption{ASO scores of models with different distillation objectives studied in Sec. \ref{ablation}. For ``\methodname~(ours) $\rightarrow$ baseline'', $\epsilon_\text{min} = 0$ indicates that \methodname~(ours) consistently outperform baseline. Here, the baseline denotes \methodname~(ours) w/o step-distillation.}
\centering 
\begin{tabular}{l|c}
\hline
{\bfseries Model} & {\bfseries ASO score ($\epsilon_\text{min}$)} \\
\hline\hline
\methodname~(ours) $\rightarrow$ baseline & 0 \\
\hline
\methodname~(ours) $\rightarrow$ CRD & 0 \\
\hline
CRD $\rightarrow$ baseline & 0.02 \\
\hline

\end{tabular}
\label{tab:st}
\end{table}

%% file: 005_conclusion.tex
\section{Conclusion}
We proposed a novel multimodal knowledge distillation method, \methodname, which leverages the strengths of learned multimodal space of the CLIP-based teacher model and large-scale parameters of the RoBERTa-based student model to perform multimodal knowledge transfer by optimizing a step-distillation objective loss. In the evaluation of two multimodal tasks, our method significantly outperforms SoTA methods up to \textbf{12.3\%} MAE score with a single modal encoder used in inference for video-level sentiment analysis, and \textbf{3.4\%} mAP for audio-visual retrieval tasks, suggesting its strengths in high efficiency-performance. Ablation studies further demonstrate the efficacy of our proposed step-distillation objective loss in improving multimodal knowledge distillation. In the next step, we will adapt meta-learning to further explore the capability of multimodal transfer learning in a few-shot setting.